\documentclass[letterpaper, 10 pt, conference]{ieeeconf}
\IEEEoverridecommandlockouts
\usepackage{bm}
\usepackage{tikz}
\usepackage{xargs}
\usepackage[hyphens]{url}
\usepackage{times}
\usepackage{array}
\usepackage{xcolor}
\usepackage{siunitx}
\usepackage{relsize}
\usepackage{amssymb}
\usepackage{amsmath}
\usepackage{flushend}
\usepackage{graphics}
\usepackage{graphicx}
\usepackage{hyperref}
\usepackage{multicol}
\usepackage{verbatim}
\usepackage{multirow}
\usepackage{booktabs}
\usepackage{pgfplots}
\usepackage{algorithm}
\usepackage{xinttools}
\usepackage{glossaries}
\usepackage{algorithmic}
\usepackage{fancyhdr}
\usetikzlibrary{arrows.meta}
\usepgfplotslibrary{groupplots}
\usepgfplotslibrary{fillbetween}
\pgfplotsset{compat=1.4}
\title{\LARGE \bf
Multi-LED Classification as Pretext For Robot Heading Estimation
}

\author{Nicholas Carlotti$^{1}$, Mirko Nava$^{1}$, and Alessandro Giusti$^{1}$%
\thanks{$^{1}$All authors are with the Dalle Molle Institute for Artificial Intelligence (IDSIA), USI-SUPSI, Lugano, 6962, Switzerland \texttt{nicholas.carlotti@idsia.ch}}%
\thanks{This work is supported by the Swiss National Science Foundation, grant number 213074.}%
}

\begin{document}

\maketitle
\thispagestyle{fancy}
\pagestyle{empty}
\fancyhead[C]{\textbf{40th Anniversary of the IEEE Conference on Robotics and Automation (ICRA@40), Rotterdam, Netherlands | September 23-26, 2024}}
\renewcommand{\headrulewidth}{0pt}


\begin{abstract}
We propose a self-supervised approach for visual robot detection and heading estimation by learning to estimate the states (\textsc{off} or \textsc{on}) of four independent robot-mounted LEDs.
Experimental results show a median image-space position error of $14$~px and relative heading MAE of \ang{17}, versus a supervised upperbound scoring $10$~px and \ang{8}, respectively.
\end{abstract}

\section{Introduction}\label{sec:intro}
Vision-based relative pose estimation of peer robots is a key capability for many multi-robot applications~\cite{dorigo2021swarm}.
While some approaches focused on handcrafted algorithms, based on circles printed on paper sheets~\cite{saska2017system} or exploiting the LEDs geometry~\cite{teixeira18}, these systems lacked adaptability to different environmental conditions.
Recent solutions train deep neural networks (DNNs) on large amounts of data.
However, collecting large quantities of images labeled with ground truth poses is expensive, especially when a variety of environments is crucial for DNN generalization~\cite{zhou2022domain}.
Common strategies to mitigate the problem are: self-supervised pre-training followed by the downstream transfer to robot localization, limiting the need for labeled samples~\cite{jing2020self};
or the use of synthetic data combined with sim-to-real approaches~\cite{zhao2020sim}.

In this work, we train a vision-based relative pose estimation DNN from large datasets, in which every image is labeled only with the true state (\textsc{off} or \textsc{on}) of each of the independent robot-mounted LEDs facing different directions.
These datasets are cheaply collected in a self-supervised manner by a pair of autonomous robots in any un-instrumented environment;
each randomly-moving robot acquires data while randomly toggling its own LEDs and communicating the LED states to its peer.
In previous work~\cite{nava2024self}, we focused on self-supervised pre-training: first, a model is trained on LED state estimation as a pretext task to learn relevant features for pose estimation; later, the pre-trained model is transferred to robot detection using pose labels.
Here, in contrast, we only assume LED state labels.

Our approach relies on two intuitions about information sources:
to accurately estimate the state of LEDs, a model has to recognize the robot on which said LEDs are fitted based solely on the information content of the image, i.e., recognizing that informative pixels for LED state estimation belong to the robot and thus developing pattern recognition for the robot's appearance.
Similarly, being able to tell the state of a particular LED depends on the robot's position and heading, as this determines which LEDs are visible;
thus, optimizing LED state estimation requires the model to recognize the robot's heading.
Consider the example of a robot with a single LED placed on the front of its chassis: assuming no label leakage, the model's best guess when said robot is visible on its back side is to predict the LED to be on or off with equal probability; instead, when the robot is seen on its front, the model must focus on the chassis and recognize the LED state.
%
%
Our chosen model is a Fully Convolutional Network (FCN)~\cite{long2015fully} that, given an image, predicts two-dimensional maps: one for the state of each LED, one representing the model's belief about the robot image-space position, and one about its heading.
We take full advantage of the inductive bias of FCNs, in which the value of a pixel in the output map depends on a limited patch of the input image, helping the model focus on the robot and ignoring other areas of the image.
We integrate the belief maps as weights in the LED state loss function such that minimizing it forces the model to detect the robot's position and orientation, i.e., we allow the model to assign large weights for visible LEDs (e.g., the front LED of the robot when its front is facing the observer in the running example) and a small weight to those that are occluded.

\section{Approach}\label{sec:approach}

\begin{figure*}[th]
    \centering
    \includegraphics[trim={0cm 0cm 0cm 0cm}, clip=true,width=1.0\textwidth]{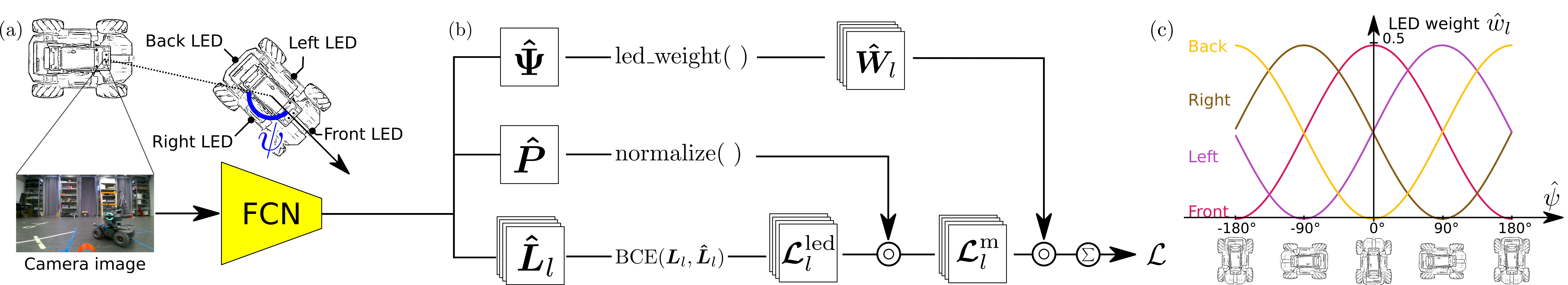}
    \vspace{-0.5mm}
    \caption{We learn robot detection and heading estimation (a) by optimizing the LED loss $\mathcal{L}$ (b) weighted by position $\bm{\hat{P}}$ and orientation $\bm{\hat{\Psi}}$ belief maps (c).}
    \label{fig:approach}
\end{figure*}

We aim to detect a robot and estimate its heading given a single monocular RGB image $\bm{I} \in \mathbb{R}^{W \times H \times 3}$, where $(u,v)$ are the image-space coordinates of the robot's frame and $\psi$ is the planar viewing direction, i.e., the angle expressed w.r.t. the detected robot's frame indicating from which direction the camera is looking at it, as shown in Figure~\ref{fig:approach}(a).
To this end, we train a deep learning model to solve the pretext task of predicting the state of the fitted robot LEDs given the camera feed.
We learn an FCN model $\bm{m_\theta}(\bm{I})$ parametrized by $\bm{\theta}$ and whose only input is the image $\bm{I}$, as shown in Figure~\ref{fig:approach}(b). 
The model produces the following two-dimensional maps: the projection map $\bm{\hat{P}} \in [0,1]^{w \times h}$, where each cell indicates the presence of a robot in the corresponding area of the image; sine and cosine maps $\bm{\hat{S}},\,\bm{\hat{C}} \in [-1, 1]^{w\times h}$ of the robot's viewing direction $\hat{\psi}$; LED state maps $\bm{\hat{L}}_l \in [0,1]^{w \times h}$ for $l=1..n$, encoded as 0 for an off LED or 1 otherwise.
%

Given these maps, we extract scalar values representing the position and viewing direction of the robot in the image: the predicted position is obtained from $\bm{\hat{P}}$ by computing its argmax.
For the heading, we compute the viewing angle map $\bm{\hat{\Psi}} = \text{atan2}(\bm{\hat{S}}, \bm{\hat{C}})$ and take the weighted average $\hat{\psi} = \text{mapsum}(\bm{\hat{\Psi}} \circ \bm{\hat{P}}_{n})$, where $\bm{\hat{P}}_{n}$ is the normalized projection map whose cells sum to one, $\circ$ denotes element-wise matrix multiplication, and $\text{mapsum}(\bm{X})\,=\,\sum_{i=1}^{w} \sum_{j=1}^{h}\bm{X}_{ij}$, giving more weight to areas in which the model detects a robot.

\subsection{Loss Function}\label{sec:approach:loss}
We train the model on LED state estimation by minimizing the binary cross-entropy (BCE) between predicted LED state and ground truth labels. We broadcast the LED ground truth labels to match the model's output size of $w \times h$, represented by $\bm{L}_l$, and define 
$\mathcal{L}^{\text{led}} = \frac{1}{n} \sum_{l=1}^{n} \text{BCE}(\bm{\hat{L}}_l, \bm{L}_l) \in \mathbb{R}^{w \times h}$.

Since most of the image does not contain information regarding the LED state, we allow our model to focus on specific areas of the image by weighting the loss with the projection map
$\mathcal{L}^{\text{m}} = \bm{\hat{P}}_{n} \circ \mathcal{L}^{\text{led}}$.
%
%
Indeed, this results in the emergent behavior of $\bm{\hat{P}}$ having high activations corresponding to the robot's image location.
This mechanism lets the model learn the detection task without position labels.

Similarly, we introduce a weighting mechanism to learn heading estimation without the respective labels.
We leverage that, based on the robot's position and heading, some LEDs are bound to be occluded from the camera while others are clearly visible; this stark difference in information access is exploited in the complete loss:
we weight $\mathcal{L}^{\text{m}}$ with the LED weighting map $\bm{\hat{W}} = \text{led\_weight}(\bm{\hat{\Psi}}) \in [0,1]^{n\times w \times h}$, where the more an LED is visible, the higher its weight.
In practice, we use cosine functions shifted according to whether the robot's body occludes an LED based on its heading $\bm{\hat{W}}_l = \cos(\bm{\hat{\Psi}} + \frac{2\cdot\pi}{n}(l - 1))$.
Further, we normalize $\bm{\hat{W}}$ cells such that weights for all LEDs sum to one. The complete loss is defined as $\mathcal{L} = \frac{1}{n}\sum_{l=1}^{n} \text{mapsum}( \text{BCE}(\bm{\hat{L}}_l, \bm{L}_l) \circ \bm{\hat{P}}_{n} \circ \bm{\hat{W}}_l)$.

\section{Experimental Setup}\label{sec:experiments}
We use two DJI S1 RoboMasters, omnidirectional ground robots equipped with four Swedish wheels and a front-facing monocular camera with a resolution of $640 \times 360$ pixels mounted on a pan-and-tilt turret.
Each robot features six LEDs: four mounted in the cardinal directions on the chassis and two laterally on the turret. For our experiments, we consider only the chassis front and back LEDs, and the left and right turret ones, as shown in Figure~\ref{fig:approach}(a), while the two others are kept off.
We let the two RoboMasters roam a $4 \times 4$ meters area inside our lab, following a random navigation policy and randomizing the state of their $n=4$ LEDs every 5 seconds.
For each robot, we collect samples containing the camera feed $\bm{I}$ at 3 FPS and the state of the LEDs $\bm{L}_l = \{0,1\}_{l=1}^{4}$. To validate the model, we also collect the robots' pose as given by an infra-red motion tracking system, denoted as $^w\bm{T}_i$ for the pose of robot $i$ w.r.t. the world reference frame.
Due to the random nature of the navigation policy, the robots are very frequently outside of each other's field of view: we report that only 22\% of samples feature images with a visible robot.
The collected data, amounting to 6 hours of recording, is split into three disjoint sets: the training set $\mathcal{T}$ with 116k samples in which we ignore the robot poses, the validation set $\mathcal{V}$ with 10k samples (2275 visible), and the testing set $\mathcal{Q}$ with 5k samples (1104 visible).

\section{Results}\label{sec:results}
We report performance metrics computed on the visible samples from $\mathcal{Q}$: the median distance between predicted and ground truth image-space robot position $E_{uv}$ and the median error for the planar viewing direction $E_{\psi}$.
In Table~\ref{tab:table}, we report metrics 
averaged over three replicas for: the \textit{Dummy Mean} model predicting the mean $u, v, \psi$ values for $\mathcal{T}$;
our proposed \textit{Pretext} model trained on $\mathcal{T}$ using only LEDs ground truth;
and the \textit{Upperbound} model sharing the same architecture as \textit{Pretext} but trained on $\mathcal{T}$ with access to ground truth labels for $u,v$ and $\psi$.
Results show that the proposed approach is effective in both localization and heading estimation: by training without labels for position and heading, \textit{Pretext} scores a median image-space position error of $14.5$px and a median heading error of \ang{17.0}, quite close to \textit{Upperbound} scoring $10.1$px and \ang{8.4} respectively; additional experimental results are shown in the accompanying \href{https://drive.switch.ch/index.php/s/9KEmsZdQvbTvakA}{\textit{video}}.

In this work, we demonstrated multi-LED state classification to learn robot detection and heading estimation, requiring only ground truth for the LED states.
Future work will focus on a more comprehensive model evaluation,
and a novel weighting mechanism for the robot's distance, enabling pose estimation without the need for pose labels.

\begin{table}[hb]
    \setlength\tabcolsep{1.2mm} 
    \renewcommand{\arraystretch}{1.0} 
    \centering
    \caption{Median 2D position $uv$ and heading $\psi$ errors on $\mathcal{Q}$.}
    \vspace{-2mm}
    \label{tab:table}
    \begin{tabular}{lrr>{\centering\arraybackslash}p{36mm}}
    \toprule
    \multirow{2}{*}{Model} & \multicolumn{1}{c}{$E_{uv}$} & \multicolumn{1}{c}{$E_{\psi}$} & {Point plot for $E_{uv}$ [px] $\leftarrow$} \\
    & [px] $\downarrow$ & [deg] $\downarrow$ & {Error bars mark 95\% conf. int.} \\
    \midrule
    Dummy Mean       & 122 & 80.2 & \multirow{1}{*}{\input{tableplot.tikz}} \\
    Pretext          &  14.5 & 17.0 & \\
    Upperbound       &  10.1 & 8.4 & \\[0.0mm]
    \bottomrule
    \end{tabular}
\end{table}

\addtolength{\textheight}{0mm}


\bibliographystyle{IEEEtran}
\bibliography{main}

\begin{thebibliography}{1}
\providecommand{\url}[1]{#1}
\csname url@rmstyle\endcsname
\providecommand{\newblock}{\relax}
\providecommand{\bibinfo}[2]{#2}
\providecommand\BIBentrySTDinterwordspacing{\spaceskip=0pt\relax}
\providecommand\BIBentryALTinterwordstretchfactor{4}
\providecommand\BIBentryALTinterwordspacing{\spaceskip=\fontdimen2\font plus
\BIBentryALTinterwordstretchfactor\fontdimen3\font minus \fontdimen4\font\relax}
\providecommand\BIBforeignlanguage[2]{{%
\expandafter\ifx\csname l@#1\endcsname\relax
\typeout{** WARNING: IEEEtran.bst: No hyphenation pattern has been}%
\typeout{** loaded for the language `#1'. Using the pattern for}%
\typeout{** the default language instead.}%
\else
\language=\csname l@#1\endcsname
\fi
#2}}

\bibitem{dorigo2021swarm}
M.~Dorigo, G.~Theraulaz, and V.~Trianni, ``Swarm robotics: Past, present, and future,'' \emph{{IEEE} Point of View}, vol. 109, no.~7, pp. 1152--1165, 2021.

\bibitem{saska2017system}
M.~Saska, T.~Baca, J.~Thomas, J.~Chudoba, L.~Preucil, T.~Krajnik, J.~Faigl, G.~Loianno, and V.~Kumar, ``System for deployment of groups of unmanned micro aerial vehicles in gps-denied environments using onboard visual relative localization,'' \emph{{Springer} Autonomous Robots}, vol.~41, no.~4, pp. 919--944, 2017.

\bibitem{teixeira18}
L.~Teixeira, F.~Maffra, M.~Moos, and M.~Chli, ``Vi-rpe: Visual-inertial relative pose estimation for aerial vehicles,'' \emph{{IEEE} Robotics and Automation Letters}, vol.~3, no.~4, pp. 2770--2777, 2018.

\bibitem{zhou2022domain}
K.~Zhou, Z.~Liu, Y.~Qiao, T.~Xiang, and C.~C. Loy, ``Domain generalization: A survey,'' \emph{{IEEE} Transactions on Pattern Analysis and Machine Intelligence}, vol.~45, no.~4, pp. 4396--4415, 2022.

\bibitem{jing2020self}
L.~Jing and Y.~Tian, ``Self-supervised visual feature learning with deep neural networks: A survey,'' \emph{{IEEE} Transactions on Pattern Analysis and Machine Intelligence}, 2020.

\bibitem{zhao2020sim}
W.~Zhao, J.~P. Queralta, and T.~Westerlund, ``Sim-to-real transfer in deep reinforcement learning for robotics: a survey,'' in \emph{{IEEE} Symposium Series on Computational Intelligence}, 2020, pp. 737--744.

\bibitem{nava2024self}
M.~Nava, N.~Carlotti, L.~Crupi, D.~Palossi, and A.~Giusti, ``Self-supervised learning of visual robot localization using led state prediction as a pretext task,'' \emph{{IEEE} Robotics and Automation Letters}, vol.~9, no.~4, pp. 3363--3370, 2024.

\bibitem{long2015fully}
J.~Long, E.~Shelhamer, and T.~Darrell, ``Fully convolutional networks for semantic segmentation,'' in \emph{{IEEE}/{CVF} Conference on Computer Vision and Pattern Recognition}, 2015, pp. 3431--3440.

\end{thebibliography}
\end{document}